\documentclass[english]{llncs}
\usepackage[T1]{fontenc}
\usepackage[latin9]{inputenc}
\usepackage{amsmath}
\usepackage{amssymb}
\usepackage{graphicx}

\makeatletter

\providecommand{\tabularnewline}{\\}

\usepackage{url}
\usepackage{tikz}
\usepackage{relsize}
\usepackage{pgfplots}
\usetikzlibrary{calc}
\usetikzlibrary{snakes}
\usetikzlibrary{decorations.pathreplacing}
\DeclareMathAlphabet{\mathcalligra}{T1}{calligra}{m}{n}

\makeatother

\usepackage{babel}
\begin{document}

\title{Using Neural Network Formalism to Solve Multiple-Instance Problems }

\author{Tomáš Pevný$^{1,2}$ and Petr Somol$^{1}$}

\institute{$^{1}$Cisco Systems, Charles Square 10, Prague 2, Czech Republic\\
$^{2}$Czech Technical University, Faculty of Electrical Engineering,
\\
Charles Square 13, Prague 2, Czech Republic}
\maketitle
\begin{abstract}
Many objects in the real world are difficult to describe by means
of a single numerical vector of a fixed length, whereas describing
them by means of a set of vectors is more natural. Therefore, \emph{Multiple
instance learning} (MIL) techniques have been constantly gaining in
importance throughout the last years. MIL formalism assumes that each
object (sample) is represented by a set (bag) of feature vectors (instances)
of fixed length, where knowledge about objects (e.g., class label)
is available on bag level but not necessarily on instance level. Many
standard tools including supervised classifiers have been already
adapted to MIL setting since the problem got formalized in the late
nineties. In this work we propose a neural network (NN) based formalism
that intuitively bridges the gap between MIL problem definition and
the vast existing knowledge-base of standard models and classifiers.
We show that the proposed NN formalism is effectively optimizable
by a back-propagation algorithm and can reveal unknown patterns inside
bags. Comparison to 14 types of classifiers from the prior art on
a set of 20 publicly available benchmark datasets confirms the advantages
and accuracy of the proposed solution. 
\end{abstract}

\section{Motivation}

The constant growth of data sizes and data complexity in real world
problems has increasingly put strain on traditional modeling and classification
techniques. Many assumptions cease to hold; it can no longer be expected
that a complete set of training data is available for training at
once, models fail to reflect information in complex data unless a
prohibitively high number of parameters is employed, availability
of class labels for all samples can not be realistically expected,
and particularly the common assumption about each sample to be represented
by a fixed-size vector seems to no longer hold in many real world
problems.

\emph{Multiple instance learning} (MIL) techniques address some of
these concerns by allowing samples to be represented by an arbitrarily
large set of fixed-sized vectors instead of a single fixed-size vector.
Any explicit ground truth information (e.g., class label) is assumed
to be available on the (higher) level of samples but not on the (lower)
level of instances. The aim is to utilize unknown patterns on instance-level
to enable sample-level modeling and decision making. Note that MIL
does not address the Representation Learning problem \cite{bengio2012RepLearn}.
Instead it aims at better utilization of information in cases when
ground truth knowledge about a dataset may be granular and available
on various levels of abstraction only.

From a practical point of view MIL promises to i) save ground truth
acquisition cost \textendash{} labels are needed on sample-level,
i.e., on higher-level(s) of abstraction only, ii) reveal patterns
on instance level based on the available sample-level ground truth
information, and eventually iii) achieve high accuracy of models through
better use of information present in data.

Despite significant progress in recent years, the current battery
of MIL tools is still burdened with compromises. The existing models
(see next Section~\ref{subsec:related} for a brief discussion) clearly
leave open space for more efficient utilization of information in
samples and for a clearer formalism to provide easily interpretable
models with higher accuracy. The goal of this paper is to provide
a clean formalism bridging the gap between the MIL problem formulation
and classification techniques of neural networks (NNs). This opens
the door to applying latest results in NNs to MIL problems.

\section{Prior art on multi-instance problem}

\label{subsec:related}The pioneering work~\cite{dietterich1997solving}
coined \emph{multiple-instance }or\emph{ multi-instance} learning
as a problem where each sample $b$ (called \emph{bag} in the following)
consists of a set of instances $x$, i.e., $b=\{x_{i}\in\mathcal{X}|i\in\{1,\ldots,|b|\}\},$
equivalently $b\in\mathcal{B}=\cup_{k>1}\{x_{i}\in\mathcal{X}|i\in\{1,\dots,k\}\}$
and each instance $x$ can be attributed a label $y_{x}\in\{-1,+1\},$
but these instance-level labels are not known even in the training
set. The sample $b$ is deemed positive if at least one of its instances
had a positive label, i.e., label of a sample $b$ is $y=\max_{x\in b}y_{x}.$
Most approaches solving this definition of MIL problem belong to \emph{instance-space
paradigm}, in which the classifier is trained on the level of individual
instances $f:\mathcal{X}\mapsto\{-1,+1\}$ and the label of the bag
$b$ is inferred as $\max_{x\in b}f(x).$ Examples of such methods
include: Diverse-density~\cite{NIPS1997_1346}, EM-DD~\cite{NIPS2001_1959},
MILBoost~\cite{NIPS2005_2926}, and MI-SVM~\cite{NIPS2002_2232}.

Later works (see reviews \cite{Amores2013,foulds2010review}) have
introduced different assumptions on relationships between labels on
the instance level and labels of bags or even dropped the notion of
instance-level labels and considered only labels on the level of bags,
i.e., it is assumed that each bag $b$ has a corresponding label $y\in\mathcal{Y},$
which is for simplicity assumed to be binary, i.e., $\mathcal{{Y}}=\{-1,+1\}$
in the following. Most approaches solving this general definition
of the problem follow either the \emph{bag-space paradigm} and define
a measure of distance (or kernel) between bags~\cite{haussler1999convolution,muandet2012learning,gartner2002multi}or
the \emph{embedded-space paradigm} and define a transformation of
the bag to a fixed-size vector~\cite{Wang:2000,Cheplygina2015b,chen:2006}. 

Prior art on neural networks for MIL problems is scarce and aimed
for \emph{instance-space paradigm. }Ref.~\cite{ramon2000multi} proposes
a smooth approximation of the maximum pooling in the last neuron as
$\frac{1}{|b|}\ln\left(\sum_{x\in b}\exp(f(x))\right),$ where $f(x):\mathcal{X}\mapsto\mathbb{R}$
is the output of the network before the pooling. Ref.~\cite{zhou2002neural}
drops the requirement on smooth pooling and uses the maximum pooling
function in the last neuron. Both approaches optimize the $L_{2}$
error function. 

Due to space limits, the above review of the prior art was brief.
The Interested reader is referred to~\cite{Amores2013,foulds2010review,carbonneau2016multiple}
for a more thorough discussion of a problem and algorithms.

\section{Neural network formalism}

\label{subsec:ProposedFormalization}The proposed neural network formalism
is intended for a general formulation of MIL problems introduced in~\cite{muandet2012learning}.
It assumes a non-empty space $\mathcal{X}$ where instances live with
a set of all probability distributions $\mathcal{P}^{\mathcal{X}}$
on $\mathcal{X}.$ Each bag corresponds to some probability distribution
$p_{b}\in\mathcal{P^{X}}$ with its instances being realizations of
random a variable with distribution $p_{b}.$ Each bag $b$ is therefore
assumed to be a realization of a random variable distributed according
to $P(p_{b},y$), where $y\in\mathcal{Y}$ is the bag label. During
the learning process each concrete bag $b$ is thus viewed as a realization
of a random variable with probability distribution $p_{b}$ that can
only be inferred from a set of instances $\{x\in b|x\sim p_{b}\}$
observed in data. The goal is to learn a discrimination function $f:\mathcal{B}\mapsto\mathcal{Y},$
where \emph{$\mathcal{B}$} is the set of all possible realizations
of distributions $p\in\mathcal{P}^{\mathcal{X}}$, i.e., $\mathcal{B}=\left\{ x_{i}|p\in\mathcal{P}^{\mathcal{X}},x_{i}\sim p,i\in\{1,\ldots l\},l\in\mathbb{N}\right\} $.
This definition includes the original used in~\cite{dietterich1997solving},
but it also includes the general case where every instance can occur
in positive and negative bags, but some instances are more frequent
in one class.

The proposed formalism is based on the \emph{embedded-space} paradigm\emph{
}representing  bag $b$ in an $m$-dimensional Euclidean space $\mathbb{R}^{m}$
through a set of mappings

\begin{equation}
\left(\phi_{1}(b),\phi_{2}(b),\ldots,\phi_{m}(b)\right)\in\mathbb{R}^{m}\label{eq:embedding}
\end{equation}
with $\phi:\mathcal{B}\mapsto\mathbb{R}.$ Many existing methods implement
embedding function as 
\begin{equation}
\phi_{i}=g\left(\left\{ k(x,\theta_{i})\right\} _{x\in b}\right),\label{eq:projection-phi}
\end{equation}
where $k:\mathcal{X}\times\mathcal{X}\mapsto\mathbb{R}_{0}^{+}$ is
a suitably chosen distance function, $g:\cup_{k=1}^{\infty}\mathbb{R}^{k}\mapsto\mathbb{R}$
is the pooling function (e.g. minimum, mean or maximum), and finally
$\Theta=\left\{ \theta_{i}\in\mathcal{X}|i\in\{1,\ldots,m\}\right\} $
is the dictionary with instances as items. Prior art methods differ
in the choice of aggregation function $g,$ distance function $k,$
and finally in the selection of dictionary items, $\Theta$. A generalization
was recently proposed in~\cite{Cheplygina2015b} defining $\phi$
using a distance function (or kernel) over the bags $k:\mathcal{B}\times\mathcal{B}\mapsto\mathbb{R}$
and dictionary $\Theta$ containing bags rather instances. This generalization
can be seen as a crude approximation of kernels over probability measures
used in~\cite{muandet2012learning}. 

\begin{figure}[t]
\begin{centering}
\includegraphics[scale=0.8]{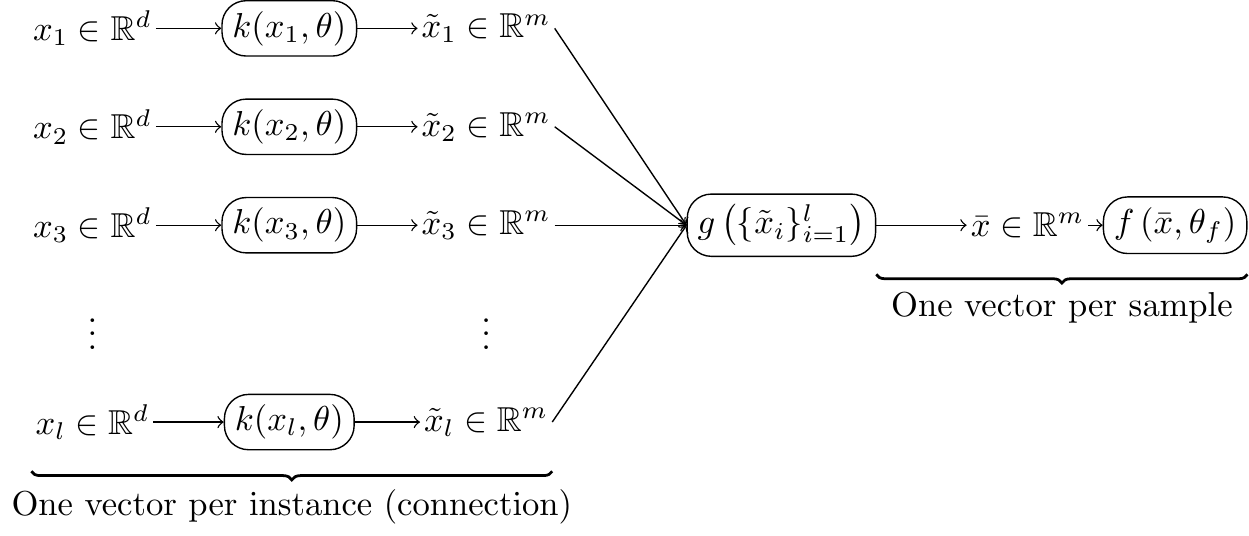}
\par\end{centering}
\caption{\label{fig:simplemil}Sketch of the neural network optimizing the
embedding in embedding-space paradigm.}
\end{figure}
The computational model defined by~(\ref{eq:embedding}) and~(\ref{eq:projection-phi})
can be viewed as a neural network sketched in Figure~\ref{fig:simplemil}.
One (or more) lower layers implement a set of distance functions $\{k(x,\theta_{i})\}_{i=1}^{m}$
(denoted in Fig.~\ref{fig:simplemil} in vector form as $k(x,\theta)$)
projecting each instance $x_{i}$ from the bag $\{x_{i}\}_{i=1}^{m}$
from the input space $\mathbb{R}^{d}$ for $\mathbb{R}^{m}.$ The
pooling layer implementing the pooling function $g$ produces a single
vector $\bar{x}$ of the same dimension $\mathbb{R}^{m}.$ Finally
subsequent layers denoted in the figure as $f(\bar{x})$ implement
the classifier that already uses a representation of the bag as a
feature vector of fixed length $m$. The biggest advantage of this
formalism is that with a right choice of pooling function $g(\cdot)$
(e.g. mean or maximum) all parameters of the embedding functions $k(x,\theta)$
can be optimized by the standard back-propagation algorithm. Therefore
embedding at the instance-level (layers before pooling) is effectively
optimized while requiring labels only on the bag-level. This mechanism
identifies parts of the instance-space $\mathcal{X}$ with the largest
differences between probability distributions generating instances
in positive and negative bags with respect to the chosen pooling function.
This is also the most differentiating feature of the proposed formalism
to most prior art, which typically optimizes embedding parameters
$\theta_{i}$ regardless of the labels.

The choice of a pooling function depends on the type of the MIL problem.
If the bag's label depends on a single instance, as it is the case
for the instance-level paradigm, then the maximum pooling function
is appropriate, since its output also depends on a single instance.
On the other hand if a bag's label depends on properties of all instances,
then the mean pooling function is appropriate, since its output depends
on all instances and therefore it characterizes the overall distribution.

Remark: the key difference of the above approach to the prior art~\cite{zhou2002neural}
is in performing pooling \emph{inside the network} as opposed to after
the last neuron or layer as in the cited reference. This difference
is key to the shift from instance-centric modeling in prior art to
bag-centric advocated here. However the proposed formalism is general
and includes~\cite{zhou2002neural} as a special case, where instances
are projected into the space of dimension one ($m=1),$ pooling function
$g$ is set to maximum, and layers after the pooling functions are
not present ($f$ is equal to identity).

\section{Experimental evaluation}

\label{sec:Experimental}The evaluation of the proposed formalism
uses publicly available datasets from a recent study of properties
of MIL problems~\cite{Cheplygina2015}, namely \emph{BrownCreeper,
CorelAfrican, CorelBeach, Elephant, Fox, Musk1, Musk2, Mutagenesis1,
Mutagenesis2, Newsgroups1, Newsgroups2, Newsgroups3, Protein, Tiger,
UCSBBreastCancer, Web1, Web2, Web3, Web4, }and \emph{WinterWren}.
The supplemental material~\cite{MILProblems} contains equal error
rate (EER) of 28 MIL classifiers (and their variants) from prior art
implemented in the MIL matlab toolbox~\cite{MIL2015} together with
the exact experimental protocol and indexes of all splits in 5-times
repeated 10-fold cross-validation. Therefore the experimental protocol
has been exactly reproduced and results from~\cite{MILProblems}
are used in the comparison to prior art.

\label{subsec:Prior}The proposed formalism has been compared to those
algorithms from prior art that has achieved the lowest error on at
least one dataset. This selection yielded 14 classifiers for 20 test
problems, which demonstrates diversity of MIL problems and difficulty
to choose suitable method. Selected algorithms include representatives
of instance-space paradigm: \emph{MIL Boost}~\cite{NIPS2005_2926},
\emph{SimpleMIL}, \emph{MI-SVM}~\cite{NIPS2002_2232} with Gaussian
and polynomial kernel, and prior art in Neural Networks (denoted \emph{prior
NN})~\cite{zhou2002neural}; bag-level paradigm: $k$-nearest neighbor
with \emph{citation} distance~\cite{Wang:2000} using 5 nearest neighbors;
and finally embedded-space paradigm: \emph{Miles}~\cite{chen:2006}
with Gaussian kernel, Bag dissimilarity~\cite{Cheplygina2015b} with
\emph{minmin}, \emph{meanmin}, \emph{meanmean},\emph{ Hausdorff},
and Earth-moving distance (EMD),\emph{ cov-coef}~\cite{MILProblems}
embedding bags by calculating covariances of all pairs of features
over the bag, and finally \emph{extremes }and \emph{mean} embedding
bags by using extreme and mean values of each feature over instances
of the bag. All embedded space paradigm methods except Miles used
a logistic regression classifier.

\begin{figure}[t]
\begin{centering}
\includegraphics[width=0.8\columnwidth]{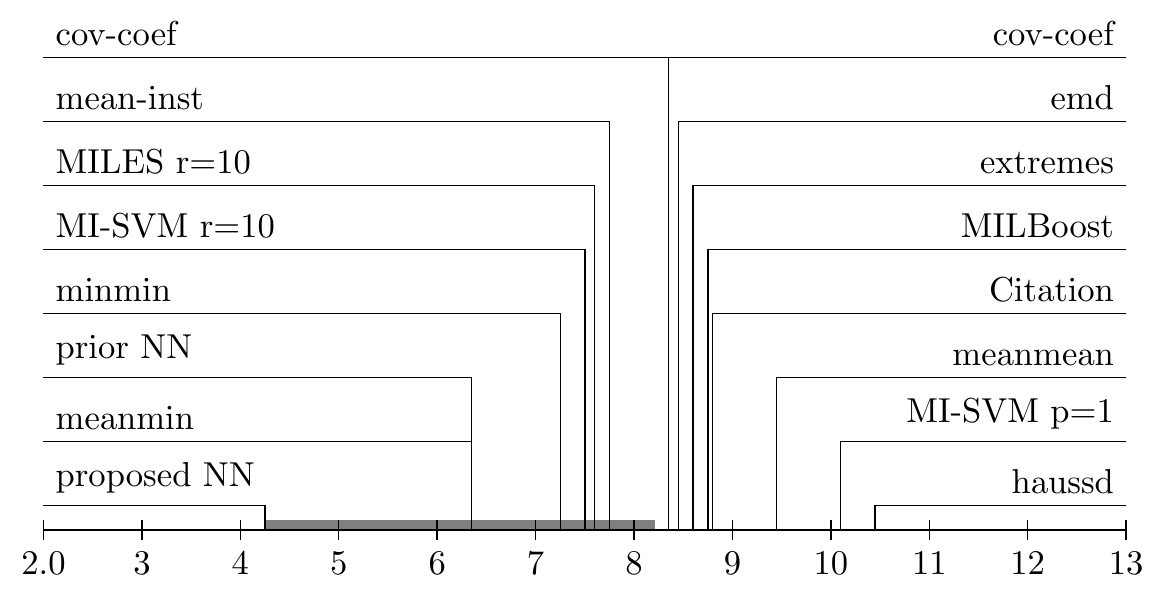}
\par\end{centering}
\caption{\label{fig:CriticalDifference}Critical difference diagram shows average
rank of each method over 20 problems. The thick black line shows the
confidence interval of corrected Bonferroni-Dunn test with significance
0.05 testing whether two classifiers have equal performance.}
\end{figure}
The proposed MIL neural network consists of a single layer of rectified
linear units (ReLu)~\cite{5459469} with transfer function $\max\{0,x\},$
followed by a mean-pooling layer and a single linear output unit.
The training minimized a hinge loss function using the Adam~\cite{kingma2014adam}
variant of stochastic gradient descend algorithm with mini-batch of
size 100, maximum of 10 000 iterations, and default settings. L1 regularization
on weights of the network was used to decrease overfitting. The topology
had two parameters \textemdash{} the number of neurons in the first
layer defining the dimension of bag representation, $m,$ and the
strength of the L1 regularization, $\lambda.$ Suitable parameters
were found by estimating equal error rates by five-fold cross-validation
(on training samples) on all combinations of $k\in\{2,4,8,12,16,20\}$
and $\lambda\in\{10^{-7},10^{-6},\ldots,10^{-3}\}$ and using the
combination achieving the lowest error. The prior art of~\cite{zhou2002neural}
was implemented and optimized exactly as the proposed approach with
the difference that the max pooling layer was \emph{after} the last
linear output unit. 

\begin{table}[t]
\centering{}%
\begin{tabular}{lccccc}
 & \multicolumn{2}{c}{Error of NN on} &  & \multicolumn{2}{c}{prior art  }\tabularnewline
\cline{2-3} \cline{5-6} 
 & training set & testing set &  & error & algorithm\tabularnewline
\hline 
BrownCreeper  & 0  & \textbf{5.0 } &  & 11.2  & MILBoost \tabularnewline
CorelAfrican  & 2.6  & \textbf{5.5 } &  & 11.2  & minmin\tabularnewline
CorelBeach  & 0.2  & \textbf{1.2 } &  & 17  & extremes\tabularnewline
Elephant  & 0  & \textbf{13.8 } &  & 16.2  & minmin\tabularnewline
Fox  & 0.4  & \textbf{33.7 } &  & 36.1  & meanmin\tabularnewline
Musk1  & 0  & 17.5  &  & \textbf{12.8}  & Citation \tabularnewline
Musk2  & 0  & \textbf{11.4 } &  & 11.8  & Hausdorff\tabularnewline
Mutagenesis1  & 7.5  & \textbf{11.8 } &  & 16.9  & cov-coef\tabularnewline
Mutagenesis2  & 14.9  & \textbf{10.0 } &  & 17.2  & emd\tabularnewline
Newsgroups1  & 0  & 42.5  &  & \textbf{18.4 } & meanmean\tabularnewline
Newsgroups2  & 0  & 35  &  & \textbf{27.5 } & prior NN\tabularnewline
Newsgroups3  & 0  & 37.5  &  & \textbf{31.2 } & meanmean\tabularnewline
Protein  & 2.5  & \textbf{7.5 } &  & 15.5  & minmin\tabularnewline
Tiger  & 0  & 20.0  &  & \textbf{19 } & MILES \tabularnewline
UCSBBreastCancer  & 0  & 25  &  & \textbf{13.6 } & MI-SVM g \tabularnewline
Web1  & 0  & 40.6  &  & \textbf{20.9 } & MILES \tabularnewline
Web2  & 0  & 28.1  &  & \textbf{7.1 } & MI-SVM p\tabularnewline
Web3  & 0  & 25  &  & \textbf{13.6 } & MI-SVM g \tabularnewline
Web4  & 0  & 18.8  &  & \textbf{1.5 } & mean-inst\tabularnewline
WinterWren  & 0  & 5.9  &  & \textbf{2.1 } & emd\tabularnewline
\hline 
\end{tabular}\caption{\label{tab:prior}Average equal error rate of the proposed NN formalism
on training and testing set and average equal error rate on the testing
set of the best prior art for the given problem. Abbreviations of
the prior art are as introduced in Section~\ref{sec:Experimental}.}
\end{table}
 Figure~\ref{fig:CriticalDifference} summarizes results in critical
difference diagram~\cite{demvsar2006statistical} showing the average
rank of each classifier over the problems together with the confidence
interval of corrected Bonferroni-Dunn test with significance 0.05
testing whether two classifiers have equal performance. The critical
diagram reveals that the classifier implemented using the proposed
neural net formalism (caption \emph{proposed NN}) achieved overall
the best performance, having the average rank 4.3. In fact, Table~\ref{tab:prior}
shows that it provides the lowest error on nine out of 20 problems.
Note that the second best, Bag dissimilarity~\cite{Cheplygina2015b}
with minmin distance and prior art in NN~\cite{zhou2002neural},
achieved the average rank 6.4 and was the best only on three and one
problems respectively.

Exact values of EER of the best algorithm from the prior art and that
of the proposed NN formalism is summarized in Table~\ref{tab:prior}.
From the results it is obvious that the proposed neural network formalism
have scored poorly on problems with a large dimension and a small
number of samples, namely Newsgroups and Web (see Table 1 of~\cite{Cheplygina2015c}
for details on the data). The neural network formalism has easily
overfit to the training data, which is supported by zero errors on
the training sets.

\section{Conclusion}

This work has presented a generalization of neural networks to multi-instance
problems. Unlike the prior art, the proposed formalism embeds samples
consisting of multiple instances into vector space, enabling subsequent
use with standard decision-making techniques. The key advantage of
the proposed solution is that it simultaneously optimizes the classifier
and the embedding. This advantage was illustrated on a set of real-world
examples, comparing results to a large number of algorithms from the
prior art. The proposed formalism seems to outperform the majority
of standard MIL methods in terms of accuracy. It should be stressed
though that results were compared to those published by authors of
survey benchmarks; not all methods in referred tests may have been
set in the best possible way. However, as many such cases would be
very computationally expensive, the proposed formalism becomes competitive
also due to its relatively modest computational complexity that does
not exceed that of a standard 3-layer neural network. The proposed
formalism opens up a variety of options for further development. A
better and possibly more automated choice of pooling functions is
one of the promising ways to improve performance on some types of
data.

\bibliographystyle{plabbrv}

\begin{thebibliography}{10}

\bibitem{Amores2013}
J.~Amores.
\newblock {Multiple instance classification: Review, taxonomy and comparative
  study}.
\newblock {\em Artificial Intelligence}, 201:81--105, 2013.

\bibitem{NIPS2002_2232}
S.~Andrews, I.~Tsochantaridis, T.~Hofmann.
\newblock Support vector machines for multiple-instance learning.
\newblock S.~Becker, S.~Thrun, K.~Obermayer, redaktorzy, {\em Advances in
  Neural Information Processing Systems 15}, strony 577--584. MIT Press, 2003.

\bibitem{bengio2012RepLearn}
Y.~Bengio, A.~Courville, P.~Vincent.
\newblock Representation learning: A review and new perspectives.
\newblock {\em arXiv preprint arXiv:1206.5538v2}, 2012.

\bibitem{carbonneau2016multiple}
M.-A. Carbonneau, V.~Cheplygina, E.~Granger, G.~Gagnon.
\newblock Multiple instance learning: A survey of problem characteristics and
  applications.
\newblock {\em arXiv preprint arXiv:1612.03365}, 2016.

\bibitem{chen:2006}
Y.~Chen, J.~Bi, J.~Z. Wang.
\newblock Miles: Multiple-instance learning via embedded instance selection.
\newblock {\em IEEE Transactions on Pattern Analysis and Machine Intelligence},
  28(12):1931--1947, Dec 2006.

\bibitem{Cheplygina2015b}
V.~Cheplygina, D.~M. Tax, M.~Loog.
\newblock Multiple instance learning with bag dissimilarities.
\newblock {\em Pattern Recognition}, 48(1):264 -- 275, 2015.

\bibitem{Cheplygina2015c}
V.~Cheplygina,  D.~M.~J. Tax.
\newblock {\em Characterizing Multiple Instance Datasets}, strony 15--27.
\newblock Springer International Publishing, Cham, 2015.

\bibitem{Cheplygina2015}
V.~Cheplygina,  D.~M.~J. Tax.
\newblock {\em Similarity-Based Pattern Recognition: Third International
  Workshop, SIMBAD 2015, Copenhagen, Denmark, October 12-14, 2015.
  Proceedings}, rozdzia/l Characterizing Multiple Instance Datasets, strony
  15--27.
\newblock Springer International Publishing, Cham, 2015.

\bibitem{MILProblems}
V.~Cheplygina, D.~M.~J. Tax, M.~Loog.
\newblock Supplemental documents to characterizing multiple instance datasets.

\bibitem{demvsar2006statistical}
J.~Dem{\v{s}}ar.
\newblock Statistical comparisons of classifiers over multiple data sets.
\newblock {\em The Journal of Machine Learning Research}, 7:1--30, 2006.

\bibitem{dietterich1997solving}
T.~G. Dietterich, R.~H. Lathrop, T.~Lozano-P{\'e}rez.
\newblock Solving the multiple instance problem with axis-parallel rectangles.
\newblock {\em Artificial intelligence}, 89(1):31--71, 1997.

\bibitem{foulds2010review}
J.~Foulds,  E.~Frank.
\newblock A review of multi-instance learning assumptions.
\newblock {\em The Knowledge Engineering Review}, 25(01):1--25, 2010.

\bibitem{gartner2002multi}
T.~G{\"a}rtner, P.~A. Flach, A.~Kowalczyk, A.~J. Smola.
\newblock Multi-instance kernels.
\newblock {\em ICML}, wolumen~2, strony 179--186, 2002.

\bibitem{haussler1999convolution}
D.~Haussler.
\newblock Convolution kernels on discrete structures.
\newblock Raport instytutowy, Universityof California at Santa Cruz, 1999.

\bibitem{5459469}
K.~Jarrett, K.~Kavukcuoglu, M.~Ranzato, Y.~LeCun.
\newblock What is the best multi-stage architecture for object recognition?
\newblock {\em Computer Vision, 2009 IEEE 12th International Conference on},
  strony 2146--2153, Sept 2009.

\bibitem{kingma2014adam}
D.~Kingma,  J.~Ba.
\newblock Adam: A method for stochastic optimization.
\newblock {\em arXiv preprint arXiv:1412.6980}, 2014.

\bibitem{NIPS1997_1346}
O.~Maron,  T.~Lozano-P\'{e}rez.
\newblock A framework for multiple-instance learning.
\newblock M.~I. Jordan, M.~J. Kearns, S.~A. Solla, redaktorzy, {\em Advances in
  Neural Information Processing Systems 10}, strony 570--576. MIT Press, 1998.

\bibitem{muandet2012learning}
K.~Muandet, K.~Fukumizu, F.~Dinuzzo, B.~Sch{\"o}lkopf.
\newblock Learning from distributions via support measure machines.
\newblock {\em Advances in neural information processing systems}, strony
  10--18, 2012.

\bibitem{ramon2000multi}
J.~Ramon,  L.~De~Raedt.
\newblock Multi instance neural networks.
\newblock 2000.

\bibitem{MIL2015}
C.~V. Tax, D.M.J.
\newblock {MIL}, a {M}atlab toolbox for multiple instance learning, Jun 2016.
\newblock version 1.2.1.

\bibitem{Wang:2000}
J.~Wang,  J.-D. Zucker.
\newblock Solving the multiple-instance problem: A lazy learning approach.
\newblock {\em Proceedings of the Seventeenth International Conference on
  Machine Learning}, ICML '00, strony 1119--1126, San Francisco, CA, USA, 2000.
  Morgan Kaufmann Publishers Inc.

\bibitem{NIPS2005_2926}
C.~Zhang, J.~C. Platt, P.~A. Viola.
\newblock Multiple instance boosting for object detection.
\newblock Y.~Weiss, B.~Sch\"{o}lkopf, J.~C. Platt, redaktorzy, {\em Advances in
  Neural Information Processing Systems 18}, strony 1417--1424. MIT Press,
  2006.

\bibitem{NIPS2001_1959}
Q.~Zhang,  S.~A. Goldman.
\newblock Em-dd: An improved multiple-instance learning technique.
\newblock T.~G. Dietterich, S.~Becker, Z.~Ghahramani, redaktorzy, {\em Advances
  in Neural Information Processing Systems 14}, strony 1073--1080. MIT Press,
  2002.

\bibitem{zhou2002neural}
Z.-h. Zhou,  M.-l. Zhang.
\newblock Neural networks for multi-instance learning.
\newblock {\em Proceedings of the international conference on intelligent
  information technology}, wolumen 182. Citeseer, 2002.

\end{thebibliography}

\end{document}